%% file: cssr.tex
\documentclass[fleqn]{article}
\usepackage{amsmath,graphicx,proceed2e,hyperref}

\input{abbreviations}

\begin{document}

\title{Blind Construction of Optimal Nonlinear Recursive Predictors for Discrete Sequences}
\author{Cosma Rohilla Shalizi \\
  Center for the Study of Complex Systems \\
  University of Michigan \\
  Ann Arbor, MI 48109\\
  \texttt{cshalizi@umich.edu} \\
  \And Kristina Lisa Shalizi \\
  Statistics Department \\
  University of Michigan \\
  Ann Arbor, MI 48109 \\
  \texttt{kshalizi@umich.edu}
}

\maketitle

\begin{abstract}
\small{We present a new method for nonlinear prediction of discrete random
  sequences under minimal structural assumptions.  We give a mathematical
  construction for optimal predictors of such processes, in the form of hidden
  Markov models.  We then describe an algorithm, CSSR (Causal-State Splitting
  Reconstruction), which approximates the ideal predictor from data.  We
  discuss the reliability of CSSR, its data requirements, and its performance
  in simulations.  Finally, we compare our approach to existing methods using
  variable-length Markov models and cross-validated hidden Markov models, and
  show theoretically and experimentally that our method delivers results
  superior to the former and at least comparable to the latter.}
\end{abstract}

\section{Introduction}

The prediction of discrete sequential data is an important problem in many
fields, including bioinformatics, neuroscience (spike trains), and nonlinear
dynamics (symbolic dynamics).  Existing prediction methods, with the exception
of variable-length Markov model (VLMM) methods, make strong assumptions about
the nature of the data-generating process.  In this paper, we present an
algorithm for the blind construction of asymptotically optimal nonlinear
predictors of discrete sequences.  These predictors take the form of minimal
sufficient statistics, naturally arranged into a hidden Markov model (HMM).  We
thus secure the many desirable features of HMMs, and hidden-state models more
generally, without having to make {\em a priori} assumptions about the
architecture of the system.  Furthermore, our method is more widely applicable
than those based on VLMMs.  We also compare our approach to the use of
cross-validation to select an HMM architecture, and find our results are at
least comparable in terms of accuracy and parsimony, and superior in terms of
speed.  For reasons of space, we omit proofs here.  These can be found in
\cite[ch.\ 5]{CRS-thesis} and \cite{AfPDiTS}.  The source code and
documentation for an implementation of CSSR are at
\url{http://bactra.org/CSSR/}.

\section{Optimal Nonlinear Predictors}

Consider a sequence of random variables $X_t$ drawn from a discrete alphabet
$\ProcessAlphabet$.  A predictive {\em statistic} is a function $\eta$ on the
past measurements $\PastT{t}$.  We want to predict the process, so we want the
statistic to summarize the information $\PastT{t}$ contains about
$\FutureT{t+1}$.  That is, we wish to maximize the mutual information between
the statistic $\eta(\PastT{t})$ and the future $\FutureT{t+1}$, i.e., in the
standard notation, maximize $I[\eta(\PastT{t});\FutureT{t+1}]$, which can be at
most $I[\PastT{t};\FutureT{t+1}]$.  A {\em sufficient} statistic is one which
reaches this level.  This implies $\eta$ is sufficient if and only if
$\eta(\past) = \eta(\pastprime)$ implies $\Prob(\FutureT{t+1}|\PastT{t} =
\past) = \Prob(\FutureT{t+1}|\PastT{t} = \pastprime)$
\cite{Kullback-info-theory-and-stats}.  Decision theory shows that optimal
prediction requires only knowledge of a sufficient statistic
\cite{Blackwell-Girshick}.  It is desirable to compress a sufficient statistic,
so as to minimize the information needed for optimal prediction.  One
sufficient statistic $\eta_1$ is smaller than another $\eta_2$ if $\eta_1$ can
be calculated from $\eta_2$.  A {\em minimal} sufficient statistic is one which
can be calculated from any other sufficient statistic.  Minimal sufficient
statistics thus are the most compact summary of the data which retains all the
predictively-relevant information.  We now construct one, following
\cite{Inferring-stat-compl} and \cite{CMPPSS}.

Say that two histories $\past$ and $\pastprime$, are equivalent when
$\Prob(\FutureT{t+1}|\PastT{t} = \past) = \Prob(\FutureT{t+1}|\PastT{t} =
\pastprime)$.  The equivalence class of $\past$ is $[\past]$.  Define the
function which maps histories to their equivalence classes:
\[
\lefteqn{\epsilon(\past) \equiv [\past]}
\]
\[
= \left\{\pastprime : \Prob(\FutureT{t+1}|\PastT{t} = \pastprime) = \Prob(\FutureT{t+1}|\PastT{t} = \past)\right\}
\]
The possible values of $\epsilon$, i.e., the equivalence classes, are known as
the ``causal states'' of the process; each corresponds to a distinct
distribution for the future. (We comment on the name ``causal state'' below.)
The state at time $t$ is $S_t = \epsilon(\PastT{t})$.  Clearly,
$\epsilon(\past)$ is a sufficient statistic.  It is also minimal, since if
$\eta$ is sufficient, then $\eta(\past) = \eta(\pastprime)$ implies
$\epsilon(\past) = \epsilon(\pastprime)$.  One can further show \cite{CMPPSS}
that $\epsilon$ is the {\em unique} minimal sufficient statistic, meaning that
any other must be isomorphic to it.

The causal states have some important properties \cite{CMPPSS}.  (1)
$\left\{S_t\right\}$ is a Markov process.  (2) The causal states are
recursively calculable; there is a function $T$ such that $S_{t+1} =
T(S_t,X_{t+1})$.  (3) One can represent the observed process $X$ as a random
function of the causal state process, i.e., there is naturally a
hidden-Markov-model representation.  (The familiar correspondence between HMMs
and state machines lets us re-phrase the second property as: the causal states
form a deterministic machine.)  We will refer to {\em causal state models}
or {\em causal state machines}.

The construction of the causal states is essentially the same as that of the
``measure-theoretic prediction process'' introduced by Frank Knight
\cite{Knight-predictive-view}, though that is framed directly in terms of the
conditional distributions.  Both can be regarded as applications of Wesley
Salmon's concept of a ``statistical relevance basis'' \cite{Salmon-1984} to
time series.  Causal states are also closely related to the ``predictive state
representations'' (PSRs) of controlled dynamical systems due to Littman, Sutton
and Singh \cite{predictive-representations-of-state}, though PSRs are not
generally minimal.  (There is currently no discovery procedure for PSRs, though
there are ways to learn the parameters of a {\em given} PSR
\cite{learning-PSRs}.)  It is not clear that the ``causal states'' really are
causal in the strong sense of e.g. \cite{Pearl-causality}, but this needs
investigation.  Meanwhile, they need a name, and ``causal states'' is less
awkward than the others.

Our algorithm for inferring the causal states from data builds on the following
observation \cite[pp.\ 842--843]{CMPPSS}.  Say that $\eta$ is {\em next-step
  sufficient} if $I[X_{t+1};\eta(\PastT{t})] = I[X_{t+1};\PastT{t}]$.  A
next-step sufficient statistic contains all the information needed for optimal
one-step-ahead prediction, but not necessarily for longer predictions.  If
$\eta$ is next-step sufficient, and it is recursively calculable, then $\eta$
is sufficient for the whole of the future. Since $\epsilon$ satisfies these
hypotheses, the minimal sufficient statistic can be found by searching among
those which are next-step sufficient and recursive.

\vspace{-1mm}
\section{Causal-State Splitting Reconstruction}

We now describe an algorithm, Causal-State Splitting Reconstruction (CSSR),
that estimates an HMM with the properties described in the last section from
sequence data.  CSSR starts by ``assuming'' the process is an independent,
identically-distributed sequence, with one causal state, and adds states when
statistical tests show that the current set of states is not sufficient.

Suppose we are given a sequence $\bar{x}$ of length $N$ from a finite alphabet
$\ProcessAlphabet$ of size $k$.  We wish to derive from this an estimate
$\EstCausalFunc$ of the the minimal sufficient statistic $\epsilon$.  We will
do this by finding a set of states $\EstCausalStateSet$, each member of which
will be a set of strings, or finite-length histories.  The function
$\EstCausalFunc$ will then map a history $\past$ to whichever state contains a
suffix of $\past$ (taking ``suffix'' in the usual string-manipulation sense).
Although each state can contain multiple suffixes, one can check \cite{AfPDiTS}
that the mapping $\EstCausalFunc$ will never be ambiguous.  (This contrasts
with the variable-length Markov models described in Sec.\ \ref{sec:VLMMs}
below, where each state contains only a single suffix.)

The \emph{null hypothesis} is that the process is Markovian on the basis of the
states in $\EstCausalStateSet$; that is,
\begin{equation}
\label{NullHyp}
\Prob(\NextObservable|\PastL = a {x}^{t-1}_{t-L+1})  = 
\Prob(\NextObservable|\EstCausalState = \EstCausalFunc(x^{t-1}_{t-L+1})) 
\end{equation}
for all $a \in \ProcessAlphabet$.  That is, adding an additional piece of
history does not change the conditional distribution for the next observation.
We can check this with a standard statistical test, such as $\chi^2$ or
Kolmogorov-Smirnov (which we used in our experiments below).  If we reject this
hypothesis, we fall back on a {\em restricted alternative hypothesis}, which is
that we have the right set of conditional distributions, but have matched them
with the wrong histories.  That is, 
\begin{equation}
\label{RestAltHyp}
\Prob(\NextObservable|\PastL = a{x}^{t-1}_{t-L+1}) = \Prob(\NextObservable|\EstCausalState = \altcausalstate)
\end{equation}
for some $\altcausalstate \in \EstCausalStateSet$, but $\altcausalstate \neq
\EstCausalFunc(x^{t-1}_{t-L+1})$.  If this hypothesis passes a statistical
test, again with size $\alpha$, then $\altcausalstate$ is the state to which we
assign the history\footnote{If more than one such state $\altcausalstate$
  exists, we chose the one for which $\ProbEst(\NextObservable|\EstCausalState
  = \altcausalstate)$ differs least, in total variation distance, from
  $\ProbEst(\NextObservable|\PastL = a {x}^{t-1}_{t-L+1})$, which is plausible
  and convenient.  However, which state we chose is irrelevant in the limit
  $N\rightarrow\infty$, so long as the difference between the distributions is
  not statistically significant.}.  Only if the restricted alternative is
itself rejected do we create a new state, with the suffix $a{x}^{t-1}_{t-L+1}$.

The algorithm itself has three phases; pseudo-code is given in Figure
\ref{figure:pseudo}.  Phase I initializes $\EstCausalStateSet$ to a single
state, which contains only the null suffix $\emptyset$.  (That is, $\emptyset$
is a suffix of any string.)  The length of the longest suffix in
$\EstCausalStateSet$ is $L$; this starts at 0.  Phase II iteratively tests the
successive versions of the null hypothesis, Eq. \ref{NullHyp}, and $L$
increases by one each iteration, until we reach some maximum length $\Lmax$.
At the end of II, $\EstCausalFunc$ is (approximately) next-step sufficient.
Phase III makes $\EstCausalFunc$ recursively calculable, by splitting the
states until they have deterministic transitions.  The last phase is not as
straightforward as it may seem.

There are standard algorithms \cite{Lewis-Papadimitriou-computation} to take a
non-deterministic finite automaton (NDFA) and produce a deterministic finite
automaton (DFA), which is equivalent in the sense of generating the same
language.  However, these algorithms do not ensure that each state of the DFA,
considered as an equivalence class of strings, is a subset of a state of the
NDFA.  In the present context, applying one of these algorithms would result in
states which mixed histories with significantly different conditional
distributions for the next symbol --- we would get a statistic which was
recursive but {\em not} next-step sufficient.  To preserve probabilistic
information while making the transitions deterministic, we proceed as follows
\cite{AfPDiTS}.  We want there to be a transition function $T(s,b)$ such that
$\EstCausalFunc(\past b) = T(s, b)$ for any $\past \in s$.  Thus, for each
state-symbol pair $s, b$, we check whether $\EstCausalFunc(\past b)$ is the
same for all $\past \in s$.  If we find a state-symbol pair where this does not
hold, we split that state into states where it does hold.  We then start
checking the state-symbol pairs all over again, since some other transitions
may have been altered.  This procedure always terminates, leaving us with a set
of states with deterministic transitions.  To do this smoothly, we must first
remove any transient states which the second phase may have created.  These
transients are never true causal states \cite{Upper-thesis}, but are sometimes
useful in filtering applications, in which case they can be straightforwardly
restored from the true, recurrent states \cite{Upper-thesis}.

$\ProbEst(X_t|\PastL = x)$ may be estimated in several ways; we have used
simple maximum likelihood.  $\ProbEst(X_t|\EstCausalState = \causalstate)$, in
turn, must be estimated, and we used the weighted average of the estimated
distributions of the histories in $\causalstate$.  When $L=0$ and the only
state contains just the null string, $\ProbEst(X_t|\EstCausalState =
\causalstate) = \ProbEst(X_t)$, the unconditional probability distribution.

\begin{figure}
\framebox{\begin{minipage}{\columnwidth}
    \textbf{Algorithm} CSSR($\ProcessAlphabet$,$\bar{x}$, $\Lmax$, $\alpha$)

\begin{minipage}{\textwidth}
  \begin{tabbing}
    I. Initialization: $L \leftarrow 0$, $\EstCausalStateSet \leftarrow \left\{\left\{\emptyset\right\}\right\}$\\
    II. \= Sufficiency:\\
    \> \textbf{while} \= $L < \Lmax$\\
    \> \> \textbf{for} \= each $\causalstate \in \EstCausalStateSet$\\
    \> \> \> estimate $\ProbEst(X_t|\EstCausalState = \causalstate)$\\
    \> \> \> \textbf{for} \= each $x \in \causalstate$\\
    \> \> \> \> \textbf{for} \= each $a \in \ProcessAlphabet$\\
    \> \> \> \> \> estimate \= $p \leftarrow \ProbEst(X_t|\PastL = ax)$\\
    \> \> \> \> \> \textsc{Test}($\EstCausalStateSet, p, ax, \causalstate, \alpha$)\\
    \> \> $L \leftarrow L+1$\\
    III. \= Recursion:\\
    \> Remove transient states from $\EstCausalStateSet$\\
    \> \texttt{recursive} $\leftarrow$ \textsc{False}\\
    \> \textbf{until} \= \texttt{recursive}\\
    \> \> \texttt{recursive} $\leftarrow$ \textsc{True}\\
    \> \> \textbf{for} \= each $\causalstate \in \EstCausalStateSet$\\
    \> \> \> \textbf{for} \= each $b \in \ProcessAlphabet$\\
    \> \> \> \> $x_0 \leftarrow$ first $x \in \causalstate$\\
    \> \> \> \> $T(s,b) \leftarrow \EstCausalFunc(x_0 b)$\\
    \> \> \> \> \textbf{for} \= each $x \in \causalstate$, $x \neq x_0$\\
    \> \> \> \> \> \textbf{if} $\EstCausalFunc(x b) \neq T(s,b)$\\
    \> \> \> \> \> \textbf{then} \= create new state $\newcausalstate \in \EstCausalStateSet$\\
    \> \> \> \> \> \> $T(\newcausalstate,b) \leftarrow \EstCausalFunc(x b)$\\
    \> \> \> \> \> \> \textbf{for} \= each \= $y \in s$ such that \\
    \> \> \> \> \> \> \> \> $\EstCausalFunc(y b) = \EstCausalFunc(x b)$\\
    \> \> \> \> \> \> \> \textsc{Move}$(y,\causalstate,\newcausalstate)$\\
    \> \> \> \> \> \> \texttt{recursive} $\leftarrow$ \textsc{False}\\

\\
\textsc{Test}\= ($\EstCausalStateSet, p, ax, \causalstate, \alpha$)\\
    \> \textbf{if} null hypothesis (Eq.\ \ref{NullHyp}) passes a test of size $\alpha$\\
    \> \textbf{then} $\causalstate \leftarrow ax \cup \causalstate$\\
    \> \textbf{else if} \= restricted alternative hypothesis (Eq.\ \ref{RestAltHyp})\\
    \> \> passes a test of size $\alpha$ for $\altcausalstate \in \EstCausalStateSet$, $\altcausalstate \neq \causalstate$\\
    \> \textbf{then} \textsc{Move}$(ax,\causalstate,\altcausalstate)$\\
    \> \textbf{else} \= create new state $\newcausalstate \in \EstCausalStateSet$\\
    \> \> \textsc{Move}$(ax,\causalstate,\newcausalstate)$\\
\\
\textsc{Move}\= ($x, \causalstate_1,\causalstate_2$)\\
    \> $\causalstate_1 \leftarrow \causalstate_1 \setminus x$\\
    \> re-estimate $\ProbEst(X_t|\EstCausalState = \causalstate_1)$\\
    \> $\causalstate_2 \leftarrow \causalstate_2 \cup x$\\
    \> re-estimate $\ProbEst(X_t|\EstCausalState = \causalstate_2)$\\
  \end{tabbing}
\end{minipage}
\end{minipage}}
\caption{\label{figure:pseudo} \small{Pseudo-code for the algorithm CSSR.
    Arguments: $\ProcessAlphabet$: discrete alphabet for the stochastic
    process; $\bar{x}$: sequence of length $N$ drawn from $\ProcessAlphabet$;
    $\Lmax$, maximum history length considered when estimating causal states;
    $\alpha$, size of the hypothesis tests, i.e., probability of falsely
    rejecting the hypothesis being tested.  Newly created states are always
    empty initially.}}
\end{figure}

\subsection{Time Complexity}

Phase I computes the relative frequency of all words in the data stream, up to
length $\Lmax + 1$.  There are several ways this can be done using just a
single pass through the data.  In our implementation, as we scan the data, we
construct a parse tree which counts the occurrences of all strings whose length
does not exceed $\Lmax + 1$.  Thereafter we need only refer to the parse tree,
not the data.  This procedure is therefore $O(N)$, and this is the only
sub-procedure whose time depends on $N$.

Phase II checks, for each suffix $ax$, whether it belongs to the same state as
its parent $x$.  Using a hash table, we can do this, along with assigning $ax$
to the appropriate state, creating the latter if need be, in constant time.
Since there are at most $u(k,\Lmax) \equiv (k^{\Lmax+1}-1)/(k-1)$ suffixes, the
time for phase II is $O(u(k,\Lmax)) = O(k^{\Lmax})$.

Phase III itself has three parts: getting the transition structure, removing
transient states, and refining the states until they have recursive
transitions.  The time to find the transition structure is at most $k
u(k,\Lmax)$.  Removing transients can be done by finding the strongly-connected
components of the state-transition graph, and then finding the recurrent part
of the connected-components graph.  Both these operations take a time
proportional to the number of nodes plus the number of edges in the
state-transition graph.  The number of nodes in the latter is at most
$u(k,\Lmax)$, since there must be at least one suffix per node, and there are
at most $k$ edges per node.  Hence transient-removal is $O(u(k,\Lmax)(k+1)) =
O(k^{\Lmax + 1} + k^{\Lmax}) = O(k^{\Lmax+1})$.  As for refining the states,
the time needed to make one refining pass is $k u(k,\Lmax)$, and the maximum
number of passes needed is $u(k,\Lmax)$, since, in the worst case, we will have
to make every suffix its own state, and do so one suffix at a time.  So the
maximum time for refinement is $O(k u^2(k,\Lmax)) = O(k^{2\Lmax + 1})$, and the
maximum time for all of phase III is $O(k^{\Lmax + 1} + k^{\Lmax + 1} +
k^{2\Lmax + 1}) = O(k^{2\Lmax + 1})$.  Note that if removing transients
consumes the maximal amount of time, then refinement cannot and vice versa.

Adding up, and dropping lower-order terms, the total time complexity for CSSR
is ${O(k^{2\Lmax+1}) + O(N)}$.  Observe that this is linear in the data size
$N$.  The high exponent in $k$ is reached only in extreme cases, when every
string spawns its own state, almost all of which are transient, etc.  In
practice, we have found CSSR to be much faster than this worst-case result
suggests\footnote{Average-case time complexity will depend on the statistical
  properties of the data source.  For instance, the number of strings of length
  $\Lmax$ is here bounded by $u(k,\Lmax) \approx k^{\Lmax}$.  But if $\Lmax$ is
  reasonably large, and the source satisfies the asymptotic equipartition
  property \cite[sec.\ 15.7]{Cover-and-Thomas}, only $\approx 2^{h\Lmax}$
  strings are produced with positive probability, where $h \leq \log{k}$ is the
  source entropy rate.}.

\section{Convergence and Performance}

We have established the convergence of CSSR on the correct set of states,
subject to suitable conditions.  The proofs, which use large deviations theory
on Markov chains, are too long to give here, so we simply state the assumptions
and the conclusions \cite{AfPDiTS}.

We make the following hypotheses.
\begin{enumerate}
\item The process is conditionally stationary.  That is, for all values of
  $\tau$, ${\Prob(\FutureT{t+1}|\PastT{t} = \past)} =
  \Prob(\FutureT{t+\tau+1}|\PastT{t+\tau} = \past)$.
\item The process has only finitely many causal states.
\item Every state contains at least one suffix of finite length.  That is,
  there is some $\Lmin$ such that every state contains a suffix of length
  $\Lmin$ or less.  This does not mean that $\Lmin$ symbols of history always
  fix the state, just that it is \emph{possible} to synchronize
  \cite{Upper-thesis} to every state after seeing no more than $\Lmin$ symbols.
\end{enumerate}

Under these assumptions, the reconstructed set of causal states ``converges in
probability'' on the true causal states, in the sense that
\[
\Prob(\exists
  \past : \CausalFunc(\past) \neq \EstCausalFunc(\past)) \rightarrow 0
\]
as $N
\rightarrow \infty$ \cite[p.\ 16]{AfPDiTS}.  We establish this by showing that
the probability of assigning a history to the wrong equivalence class goes to
zero as $N\rightarrow\infty$.  More exactly, for a pair of histories $\past,
\pastprime$, define the events
\begin{eqnarray*}
E(\past,\pastprime) &\equiv & (\EstCausalFunc(\past) = \EstCausalFunc(\pastprime)) \wedge (\CausalFunc(\past) \neq
\CausalFunc(\pastprime)), \\
F(\past,\pastprime) & \equiv & (\EstCausalFunc(\past) \neq \EstCausalFunc(\pastprime)) \wedge (\CausalFunc(\past) = \CausalFunc(\pastprime)) ~.
\end{eqnarray*}
Then \cite[pp.\ 15--16]{AfPDiTS}
\[
\forall \past,\pastprime, ~ \Prob(E(\past,\pastprime) \cup F(\past,\pastprime)) \rightarrow 0 ~.
\]
To establish this fact in its turn, we use large deviation theory for Markov
chains to show that the empirical conditional distribution for each history
converges on its true value exponentially quickly \cite[pp.\ 12--13]{AfPDiTS},
and consequently the probability that any of our estimated conditional
distributions differs significantly from its true value goes to zero
exponentially in $N$ \cite[pp.\ 13--15]{AfPDiTS}.  The test size $\alpha$ does
not affect this convergence, becoming irrelevant in the limit of large $N$.
(With finite $N$, of course, $\alpha$ influences our risk of making states
simply on the basis of sampling fluctuations.)  Under some further assumptions,
$\Prob(\EstCausalFunc \neq \CausalFunc)$ actually goes to zero exponentially in
$N$, and then, by the Borel-Cantelli Lemma, one has the wrong structure only
finitely often before converging on the true causal states.

\begin{figure}[t]
\begin{center}
\resizebox{\columnwidth}{!}{\includegraphics{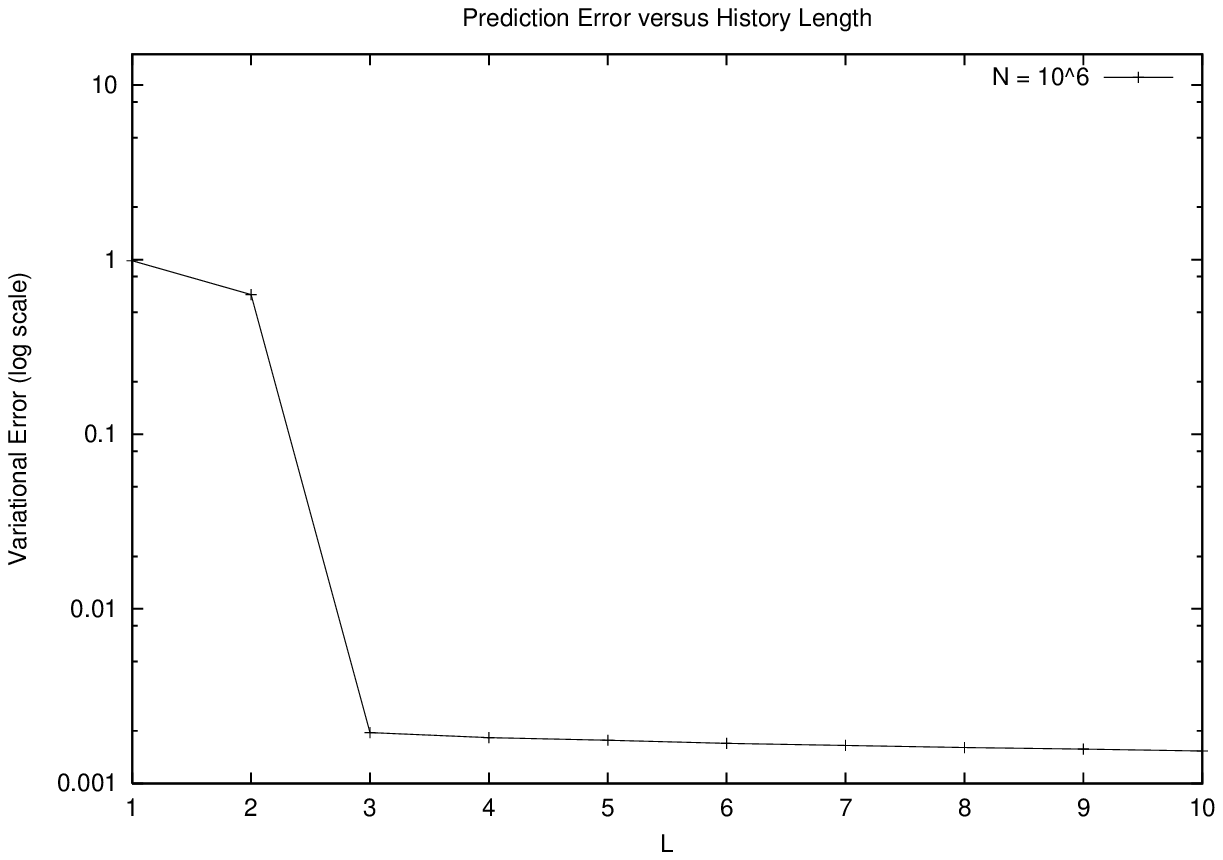}}
\resizebox{\columnwidth}{!}{\includegraphics{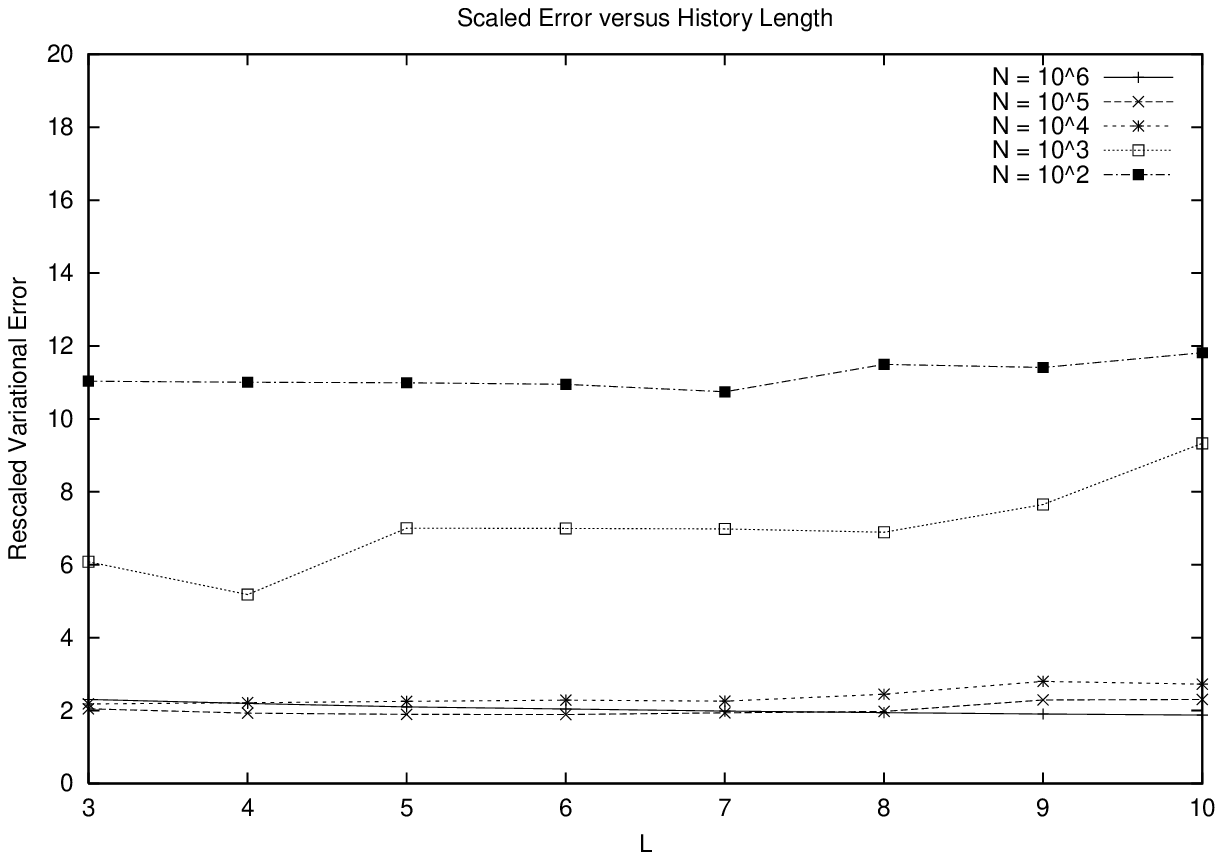}}
\end{center}
\caption{\small{Prediction error as a function of $\Lmax$ and $N$ for the even
    process.  Error is the total-variation distance between the actual
    distribution over words of length 10, and that predicted by the inferred
    states.  Top panel: error (log scale) as a function of $\Lmax$.  Bottom:
    error times $\sqrt{N}$ (linear scale).  Here $3 \leq \Lmax \leq 10$, since
    if $\Lmax < 3$ CSSR cannot find the correct states.  With this $\alpha$,
    CSSR never gets the states right for $N = {10}^{2}$, and only sporadically
    for $N = {10}^3$, so those lines are not on the scaling curve.}}
\label{figure:EP-error}
\label{figure:EP-error-scaling}
\end{figure}

If the states are correct, then another large-deviation argument
\cite[sec.\ 4.3]{AfPDiTS} gives us a handle on the expected prediction error.
Since the forecasts made by our predictors are distributional, error should be
measured as a divergence between the predicted distribution and the true one.
Using the total variation metric\footnote{The total variation or $L_1$ distance
  between two measures $P$ and $Q$ over a discrete space $\ProcessAlphabet$ is
  $d(P,Q) \equiv \sum_{a \in \ProcessAlphabet}{\left|P(a) - Q(a)\right|}$.
  Scheffe's identity \cite{Devroye-Lugosi-combinatorial} asserts that $d(P,Q) =
  2\sup_{\mathrm{A} \subseteq \ProcessAlphabet}{\left|P(\mathrm{A}) -
    Q(\mathrm{A})\right|}$, and consequently $0 \leq d(P,Q) \leq 2$.}  as our
divergence measure, error goes down as $N^{-1/2}$.  This is illustrated in
Figure \ref{figure:EP-error-scaling}.

\begin{figure}[t]
\begin{center}
\resizebox{\columnwidth}{!}{\includegraphics{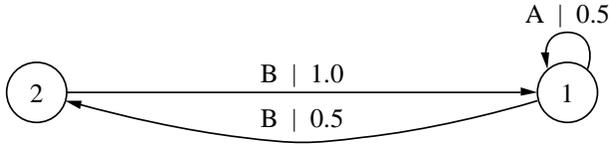}}
\end{center}
\caption{\small{The Even Process.  Transition labels show the symbol emitted,
    and the probability of making the transition.}}
\label{figure:EvenProcess}
\end{figure}

\begin{figure}[t]
\begin{center}
\resizebox{\columnwidth}{!}{\includegraphics{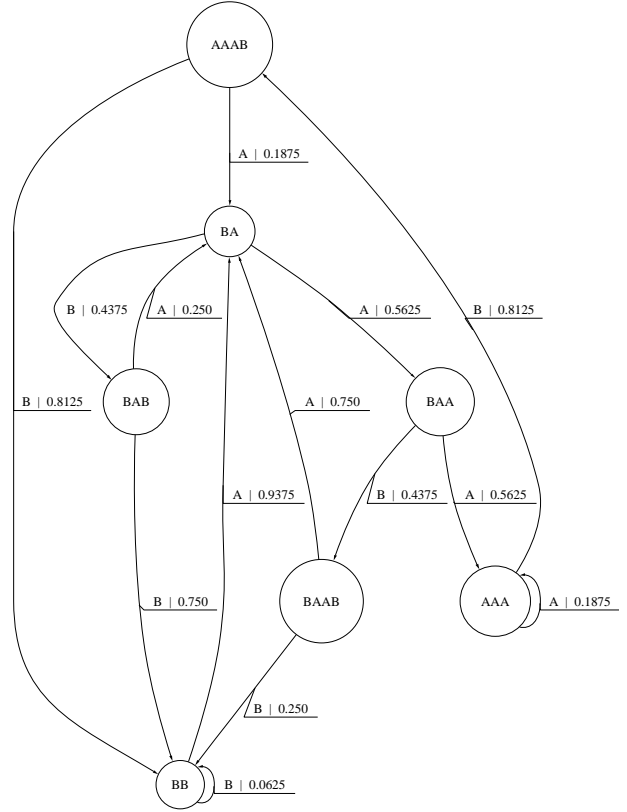}}
\end{center}
\caption{\small{A seven-state process, used in \cite{Feldman-Hanna-on-Foulkes}
    to study human sequence prediction.  Here each state is defined by a single
    suffix, as indicated.  All transition probabilities are multiples of
    $1/16$.}}
\label{figure:Foulkes-process}
\end{figure}

Of the three assumptions in our convergence proof, the only one which
affects the parameters of the algorithm is the third, that there is an integer
$\Lmin$ such that each of the true causal states contains at least one suffix
of length $\Lmin$ or less.  $\Lmin$ is thus a characteristic of the underlying
process, not CSSR.  If $\Lmax < \Lmin$, not only does the proof of convergence
fail, there is no way to obtain the true states.  For periodic processes
$\Lmin$ is equal to the period; there are no general results for other kinds of
processes.

Rather than guessing $\Lmin$, one might use the largest feasible $\Lmax$, but
this is limited by the quantity of data \cite{Marton-Shields}.  Let $L(N)$ be
the the maximum $L$ we can use when we have $N$ data-points.  If the observed
process has the weak Bernoulli property (which random functions of irreducible
Markov chains do), and an entropy rate of $h$, then a sufficient condition for
the convergence of sequence probability estimates is that $L(N) \leq \log{N}/(h
+ \varepsilon)$, for some positive $\varepsilon$.  If $L(N) \geq \log{N}/h$,
probability estimates over length $L$ words do not converge.  We must know $h$
to use this result, but $\log{k} \geq h$, so using $\log{k}$ in those formulas
gives conservative estimates.  For a given process and data-set, it is of
course possible that $L(N) < \Lmin$, in which case we simply haven't enough
data to reconstruct the true states.

We have tested CSSR on a variety of real and simulated data sources, and here
report two simulated examples, both binary-valued: the ``even process,''
illustrated in Figure \ref{figure:EvenProcess}, and a seven-state process used
in experimental studies of human sequence prediction
\cite{Feldman-Hanna-on-Foulkes}, illustrated in figure
\ref{figure:Foulkes-process}.  (The results of applying CSSR to neuronal spike
trains will be reported elsewhere.)

For the even process, the system can start in either state 1 or state 2.  When
in state 1 it is equally likely to emit an A, staying in 1, or emit a B, moving
to 2.  In 2 it always emits a B and moves to 1.  This is an HMM, but it is not
equivalent to any finite-order Markov chain (see below).  Figures
\ref{figure:EP-number-states} and \ref{figure:EP-error} illustrate the ability
of CSSR to get the correct number of causal states, the correct transition
structure, and the correct distributions.  Figure \ref{figure:EP-error} also
shows the asymptotic scaling of the error with $N$.  Curves average over 30
independent trials at each $N$; $\alpha$ is fixed to ${10}^{-3}$.  Results for
the seven-state process of Figure \ref{figure:Foulkes-process} are given
in Section \ref{sec:cross-validation}.

\begin{figure}[t]
\begin{center}
\resizebox{\columnwidth}{!}{\includegraphics{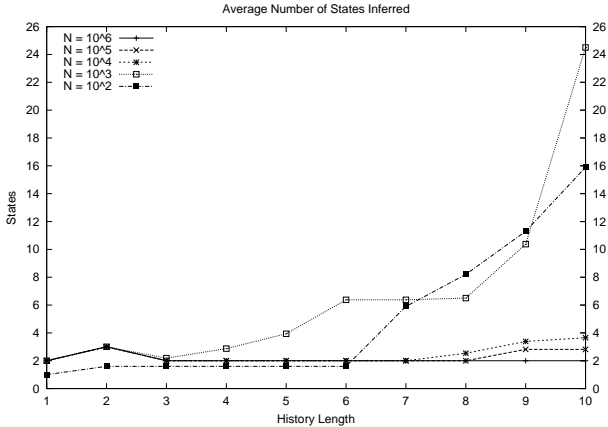}}
\end{center}
\caption{\small{Number of states inferred versus $\Lmax$ and $N$ for the even
    process.  The true number of causal states is 2.}}
\label{figure:EP-number-states}
\end{figure}

\section{Comparison with Previous Methods}

\subsection{Variable-Length Markov Models}
\label{sec:VLMMs}

The ``context'' algorithm of Rissanen \cite{Rissanen-1983} and its descendants
\cite{Buhlmann-Wyner,Willems-Shtarkov-Tjalkens-CTW,Tino-Dorffner-fractal,%
  Kennel-Mees-context-tree-modeling} construct ``variable-length Markov
models'' (VLMMs) from sequence data.  They find a set of \emph{contexts} such
that, given the context, the past of the sequence and its next symbol are
conditionally independent.  Contexts are taken to be suffixes of the history,
and the algorithms work by examining increasingly long histories, creating new
contexts by splitting existing ones into longer suffixes when thresholds of
error are exceeded \cite{Ron-Singer-Tishby-amnesia}.  (This means that contexts
can be arranged in a tree, so these are also called ``context tree'' or
``probabilistic suffix tree'' \cite{Ron-Singer-Tishby-amnesia} algorithms.)

Causal state reconstruction has an important advantage over VLMM methods.  Each
state in a VLMM is represented by a single suffix, and consists of all and only
the histories ending in that suffix.  For many processes, the causal states
contain multiple suffixes.  In these cases, multiple ``contexts'' are needed to
represent a single causal state, so VLMMs are generally more complicated than
the HMMs we build.  The causal state model is the same as the minimal VLMM if
and only if every causal state contains a {\em single} suffix.  This is the
case for the process in Fig.\ \ref{figure:Foulkes-process}, where CSSR and VLMM
methods will give the same results.

Recall the even process of the last section.  Any history ending in A, or in an
A followed by an even number of B's, belongs to state 1.  Any history
terminated by an A followed by an odd number of B's, belongs to 2.  Clearly 1
and 2 both contain infinitely many suffixes, and so correspond to an infinite
number of contexts.  VLMMs are simply incapable of capturing this structure.
If we let $\Lmax$ grow, a VLMM algorithm will increase the number of contexts
it finds without bound, but cannot achieve the same combination of predictive
power and model simplicity as causal state reconstruction (as illustrated by
Figures \ref{figure:EP-number-states} and \ref{figure:EP-error}).  Note, too,
that the causal states for the even process have finite representations, even
though they contain infinitely many suffixes.

The even process is one of the \emph{strictly sofic processes}
\cite{Weiss-1973,Badii-Politi}, which can be described by finite state models,
but are not Markov chains of any finite order\footnote{More exactly, each
  history $\past$ has a \emph{follower set} of futures $\future$ which can
  succeed it.  A process is \emph{sofic} if it has only a finite number of
  distinct follower sets, and strictly sofic if it is sofic and has an infinite
  number of irreducible forbidden words.}.  Just as VLMMs cannot handle the
even process, they cannot handle any strictly sofic process, even though those
are just regular languages.  Causal states cannot provide a finite
representation of every stochastic regular language \cite{Upper-thesis}, but
the class they capture strictly includes those captured by VLMMs.

\subsection{Cross-Validation}
\label{sec:cross-validation}

A standard heuristic for finding the right HMM architecture is cross-validation
\cite{tEoSL}.  One picks multiple candidate architectures, training each one
using the expectation-maximization (EM) algorithm, and then compares their
performance on fresh test data, selecting the one with the smallest
out-of-sample error.

To compare the performance of CSSR against this baseline, we started with
fully-connected HMMs with $M$ states, $M = 1$ to 10.  We trained these, using
the EM algorithm, on $N$ data-points from the even process.  Our test data
consisted of another $N$ data-points from an independent realization, and we
selected HMMs based on the log-likelihood they assigned to the test data.
Following common practice, the initial HMM parameters fed to the EM algorithm
were those for fully-connected models, i.e., every state could transition to
every other state, and every state could emit every symbol.  We then
calculated, for each cross-validated HMM, the total variation distance between
the distributions it and the even process generated over sequences of length
10.  (Because cross-validation is so computationally intensive, we have only
compared up to length $N={10}^4$.)  Table \ref{table:HMMs-suck} compares this
error measure for the cross-validated HMMs and for the reconstructed causal
state models.  It also indicates the number of states selected by
cross-validation, which is consistently higher than the number needed by CSSR.
Table \ref{table:HMMs-suck-harder} gives the results of a completely parallel
procedure applied to the seven-state process.

\begin{table}[t]
\resizebox{\columnwidth}{!}{\begin{tabular}{l|ll|ll}
$N$ & $d_{CV}$ & $d_{CSSR}$ & $\hat{s}_{CV}$ & $\hat{s}_{CSSR}$ \\
\hline
${10}^2$ &  $1.27 \pm 0.23$ & $1.10 \pm 0.23$ & $6.6 \pm 1.5$ & $1.6 \pm 1.0$\\
${10}^3$ &  $1.25 \pm 0.41$ & $0.19 \pm 0.23$ & $5.6 \pm 1.7$ & $2.2 \pm 0.1$\\
${10}^4$ &  $1.15 \pm 0.02$ & $0.02 \pm 0.02$ & $2.0 \pm 0  $ & $2.0 \pm 0$\\
\end{tabular}}
\caption{\label{table:HMMs-suck} \small{Comparison of the performance of HMM
    cross-validation to CSSR on the even process. $d_{CV}$, total-variation
    distance between cross-validated HMM and the even process; $d_{CSSR}$,
    distance between reconstructed causal state model and the even process;
    $\hat{s}_{CV}$, number of states in cross-validated HMM; $\hat{s}_{CSSR}$,
    number of states in reconstructed model.  In all cases the numbers given
    are the means over multiple independent trials, plus or minus one standard
    deviation.  Recall that $0 \leq d \leq 2$, and that the minimal number of
    states needed is 2.}}
\end{table}

CSSR, like the VLMM methods, is a constructive approach.  Cross-validation is
not constructive but selective.  In our case, starting with fully-connected
models (which, again, is a standard heuristic), cross-validated
expectation-maximization {\em never} selected a model whose structure
corresponded to the minimal sufficient statistic of the data-generating
process.  In both cases, the generalization of HMMs with more states worsened
as the data-length grew, so cross-validation increasingly favored small HMMs
which, while bad predictors, at least did not over-fit.  Had models with the
correct structure been in the initial population of candidates, they doubtless
would have done quite well, and the gap in predictive performance between CSSR
and cross-validation would be much smaller.  Even when we have such prior
architectural knowledge, CSSR will typically be faster than cross-validated EM,
which involves performing nonlinear optimization on multiple model structures.

\begin{table}[t]
\resizebox{\columnwidth}{!}{\begin{tabular}{l|ll|ll}
$N$ & $d_{CV}$ & $d_{CSSR}$ & $\hat{s}_{CV}$ & $\hat{s}_{CSSR}$ \\
\hline
${10}^2$ &  $ 1.41 \pm 0.23$ & $ 0.70 \pm 0.12 $ & $ 4.5 \pm 2.1$ & $ 5.1 \pm 1.5$\\
${10}^3$ &  $ 1.40 \pm 0.17$ & $ 0.21 \pm 0.06 $ & $ 5.8 \pm 2.7$ & $ 6.6 \pm 0.8$\\
${10}^4$ &  $ 1.40 \pm 0.11$ & $ 0.06 \pm 0.01 $ & $ 2.3 \pm 0.7$ & $ 7.2 \pm 0.6$\\
\end{tabular}}
\caption{\label{table:HMMs-suck-harder} \small{Comparison of the performance of
    HMM cross-validation to CSSR on the seven-state process. Variables are as
    in \ref{table:HMMs-suck}.}}
\end{table}

\section{Conclusion}

We have described an algorithm, CSSR, for the unsupervised construction of
optimal nonlinear predictors of discrete sequences.  The predictors take the
form of minimal sufficient statistics, arranged naturally into a hidden Markov
model.  CSSR's time complexity is linear in the data size.  It reliably infers
the statistical structure of processes with finitely many causal states.
CSSR's predictive performance is at least comparable to cross-validated
expectation-maximization, but it is constructive and faster; and the class of
processes it can represent is strictly larger than those of competing
constructive methods, such as variable-length Markov models.

Two directions for future work suggest themselves.  (1) CSSR does not {\em
  require} prior knowledge about system dynamics, but by the same token cannot
exploit such knowledge when it exists.  One way around this would be to
initialize the algorithm with a non-trivial partition of histories, reflecting
a guess about which patterns are dynamically important, and let CSSR revise
that partition the way it does now.  It would be interesting to know when CSSR
could correct an erroneous initial partition.  (2) HMMs are models of dynamical
systems without inputs.  Partially-observable Markov decision processes
(POMDPs) model systems with inputs, i.e., controlled dynamical systems.  The
causal state theory we have used generalizes naturally to this setting
\cite[ch.\ 7]{CRS-thesis}, where it is especially closely connected to PSRs.
Suffix-tree methods can induce POMDPs from data
\cite{Ron-Singer-Tishby-amnesia,McCallum-utile-distinctions,%
  Jaeger-operator-models}, and we believe CSSR can be adapted to reconstruct
POMDPs.  This would provide a discovery procedure for PSRs.

\section*{Acknowledgments}

\small{For support, we thank the Santa Fe Institute (under grants from Intel,
  the NSF and the MacArthur Foundation, and DARPA cooperative agreement
  F30602-00-2-0583), the NSF Research Experience for Undergraduates Program
  (KLS), and the James S. McDonnell Foundation (CRS).  Thanks to D. J. Albers,
  S. S. Baveja, P.-M. Binder, J. P. Crutchfield, D. P. Feldman, R. Haslinger,
  M. Jones, C. Moore, S. Page, A. J. Palmer, A. Ray, D. E. Smith, D. Varn and
  anonymous referees for suggestions, R. Haslinger for reading the MS.,
  J. Lindsey and S. Iacus for \texttt{R} help, E.\ van Nimwegen for providing a
  preprint of \cite{Bussemaker-et-al-genome-dictionary} and suggesting that
  something similar might infer causal states, K. Kedi for moral support in
  programming and writing, and G. Richardson for initiating our collaboration.}

\small{\bibliography{locusts}}
\bibliographystyle{unsrt-abbrv}

\end{document}

%% file: abbreviations.tex
\newcommand{\past}	{ x^{-}}
\newcommand{\PastT}[1]{X^{#1}_{-\infty}}
\newcommand{\FutureT}[1]{X^{\infty}_{#1}}
\newcommand{\pastprime}{y^{-}}

\newcommand{\future}	{ x^{+}}

\newcommand{\PastL}	{ X^{t-1}_{t-L}}

\newcommand{\causalstate}	{ s }

\newcommand{\NextObservable} {X_{t}}
\newcommand{\Prob}	{ {\rm P} }

\newcommand{\EstCausalState}	{\hat{S}}
\newcommand{\EstCausalStateSet}	{\Sigma}
\newcommand{\EstCausalFunc}	{\hat{\epsilon}}
\newcommand{\CausalFunc}	{\epsilon}

\newcommand{\ProcessAlphabet}	{\mathcal{A}}
\newcommand{\ProbEst}			{ {\widehat{\Prob}}}

\newcommand{\Lmax}{L_{\mathrm{max}}}
\newcommand{\Lmin}{\Lambda}
\newcommand{\altcausalstate}{\causalstate^{*}}
\newcommand{\newcausalstate}{\causalstate^{\prime}}